\pdfoutput=1

\documentclass[11pt]{article}

\usepackage{acl}

\usepackage{times}
\usepackage{latexsym}
\usepackage{booktabs}
\usepackage{colortbl}
\usepackage{pythonhighlight}
\usepackage{xspace}
\usepackage{graphicx}
\usepackage{pifont}
\definecolor{lp}{HTML}{CBC3E3}

\usepackage[T1]{fontenc}

\usepackage[utf8]{inputenc}

\usepackage{microtype}

\usepackage{inconsolata}

\usepackage{graphicx}

\usepackage{multirow}
\usepackage{listings}
\usepackage{amssymb}
\usepackage{amsmath}

\newcommand\blfootnote[1]{%
  \begingroup
  \renewcommand\thefootnote{}\footnote{#1}%
  \addtocounter{footnote}{-1}%
  \endgroup
}

\title{Open Ko-LLM Leaderboard2: \\ Bridging Foundational and Practical Evaluation for Korean LLMs}

\author{Hyeonwoo Kim$^{1}$, Dahyun Kim$^{2}$, Jihoo Kim$^{1}$ \\ {\bf \large Sukyung Lee$^{1}$, Yungi Kim$^{3}$, Chanjun Park$^{4\dagger}$}\\
\\
  $^{1}$Upstage AI, $^{2}$Twelve Labs, $^{3}$Liner, $^{4}$Korea University \\
  \texttt{\normalsize\{choco\_9966, jerry, sukyung\}@upstage.ai}, \texttt{\normalsize{kian.kim}@twelvelabs.io}\\ \texttt{\normalsize{eddie}@linercorp.com} \\
  \texttt{\normalsize{bcj1210}@korea.ac.kr}
  }

\begin{document}
\maketitle
\begin{abstract}
\blfootnote{$^\dagger$ Corresponding Author}
The Open Ko-LLM Leaderboard has been instrumental in benchmarking Korean Large Language Models (LLMs), yet it has certain limitations. Notably, the disconnect between quantitative improvements on the overly academic leaderboard benchmarks and the qualitative impact of the models should be addressed. Furthermore, the benchmark suite is largely composed of translated versions of their English counterparts, which may not fully capture the intricacies of the Korean language. To address these issues, we propose Open Ko-LLM Leaderboard2, an improved version of the earlier Open Ko-LLM Leaderboard. The original benchmarks are entirely replaced with new tasks that are more closely aligned with real-world capabilities. Additionally, four new native Korean benchmarks are introduced to better reflect the distinct characteristics of the Korean language. Through these refinements, Open Ko-LLM Leaderboard2 seeks to provide a more meaningful evaluation for advancing Korean LLMs.
\end{abstract}

\section{Introduction}
The Open Ko-LLM Leaderboard was originally established as a critical evaluation platform to benchmark Korean-specific Large Language Models (LLMs)~\cite{park2024open,park2024understanding}. Its motivation stemmed from the growing need to adapt existing English-centric benchmarks to Korean, thereby fostering the development of language models that can effectively handle the complexities of Korean syntax and semantics. However, the leaderboard has faced significant limitations over time.

For instance, as improvements in benchmark scores no longer translated to real-world advancements due to the overly academic nature of the benchmark suite, submission rates decreased as the leaderboard results were not as meaningful as before. 
The benchmark suite need tasks that correlate more with real-world performance.
Further, the leaderboard’s tasks, primarily configured by translating English counterparts, do not sufficiently capture the nuances of the Korean language. In fact, although the leaderboard was designed for Korean LLMs, only one of the five benchmarks, Ko-CommonGen v2, was specifically tailored for Korean, highlighting a gap in its linguistic specificity.

To address these challenges, we propose the Open Ko-LLM Leaderboard2. This next-generation framework replaces the previous benchmarks with a suite of tasks focusing on Korean linguistic nuances and real-world applications. Notably, the introduction of KorNAT benchmarks~\cite{lee2024kornat} and practical, real-world evaluations like Ko-IFEval~\cite{zhou2023instruction} and Ko-GPQA~\cite{rein2023gpqa} ensures the leaderboard’s continued relevance. Furthermore, the shift toward fine-tuned models aligns with industry trends, enabling a more meaningful assessment of task-specific performance in Korean LLMs~\cite{peng2024survey,guo2023evaluating}.

\section{Open Ko-LLM Leaderboard Season 1}
The Open Ko-LLM Leaderboard (Season 1)~\cite{park2024open,park2024understanding} was established to provide a comprehensive evaluation framework for Korean-specific Large Language Models (LLMs). Its development was driven by two primary motivations: (i) ensuring alignment with the English Open LLM Leaderboard to facilitate consistent and comparable evaluations across global and Korean LLMs, and (ii) utilizing private test sets to prevent data contamination and ensure rigorous evaluation across a variety of models.

The evaluation relied on the Ko-H5 benchmark, which consisted of five tasks: Ko-ARC~\cite{clark2018think}, Ko-HellaSwag~\cite{zellers2019hellaswag}, Ko-MMLU~\cite{hendrycks2020measuring}, Ko-TruthfulQA~\cite{lin2021truthfulqa}, and Ko-CommonGen v2~\cite{seo2024kocommongen}. While these tasks provided a foundational assessment of Korean LLMs, four of the five benchmarks were direct translations from English datasets, limiting their linguistic specificity. Only Ko-CommonGen v2 was developed with a focus on Korean, underscoring the need for more Korean-centric benchmarks in future iterations.

\section{Open Ko-LLM Leaderboard2}
\subsection{Task Overview}
The Open Ko-LLM Leaderboard2 introduces a comprehensive overhaul of its evaluation framework by replacing all previous benchmarks with nine newly designed tasks. These tasks assess a wide range of linguistic and practical capabilities essential for testing Korean LLMs in both academic and real-world settings. 

The newly added benchmarks are as follows. \textit{Ko-GPQA (Diamond)}~\cite{rein2023gpqa}, a general-purpose question-answering task that evaluates deep reasoning in the Korean context. \textit{Ko-WinoGrande}~\cite{sakaguchi2021winogrande} focuses on commonsense reasoning by challenging models to resolve ambiguities in everyday Korean scenarios. \textit{Ko-GSM8K}~\cite{cobbe2021training} assesses mathematical reasoning, requiring models to solve complex arithmetic and word problems. \textit{Ko-EQ-Bench}~\cite{paech2023eq} tests emotional intelligence by evaluating the model’s ability to generate contextually appropriate responses in emotionally charged conversations. \textit{Ko-IFEval}~\cite{zhou2023instruction} examines instruction-following skills, gauging how well models can interpret and execute complex Korean instructions. \textit{KorNAT-Knowledge}~\cite{lee2024kornat}, a newly introduced benchmark, tests factual recall and application in Korean-specific contexts. \textit{KorNAT-Social-Value}~\cite{lee2024kornat} evaluates models on their understanding of social norms and values that are unique to Korean culture. \textit{Ko-Harmlessness}~\cite{lee2024kornat} measures the model's capacity to produce safe and non-toxic responses in sensitive scenarios, while \textit{Ko-Helpfulness}~\cite{lee2024kornat} focuses on the model's ability to provide relevant and practical information across a variety of real-world situations.

\subsection{Task Motivation}
The selection of the newly added benchmarks was guided by considerations of cost-efficiency, task diversity, and practical applicability, resulting in a comprehensive yet scalable evaluation framework for Korean LLMs.

First, cost-efficiency was prioritized by adopting GPT-free automated evaluation methods, which significantly reduced costs. The dataset sizes were optimized to balance evaluation depth and computational efficiency, minimizing time and resource requirements while maintaining reliability. This approach ensures a practical and accessible evaluation process.

Second, task diversity was central to the benchmark design, covering both general LLM capabilities, such as reasoning (\textit{Ko-WinoGrande}, \textit{Ko-GPQA}, \textit{Ko-GSM8K}), instruction-following (\textit{Ko-IFEval}), and emotional intelligence (\textit{Ko-EQ-Bench}), and Korea-specific elements like cultural knowledge (\textit{KorNAT-Knowledge}) and social values (\textit{KorNAT-Social-Value}). Furthermore, tasks on harmlessness (\textit{Ko-Harmlessness}) and helpfulness (\textit{Ko-Helpfulness}) ensure safe and practical results in real-world scenarios.

Lastly, practical considerations shaped the selection of the benchmark. The evaluation framework was inspired by the Open LLM Leaderboard, ensuring consistency with established evaluation standards. The task configurations were calibrated to match the submission volumes, guaranteeing scalability and feasibility.

Overall, the chosen benchmarks achieve a thoughtful balance of evaluation rigor, efficiency, and relevance, providing a reliable platform to assess the diverse capabilities of Korean LLMs.

\subsection{Dataset Sizes}
Each of the nine benchmarks in the Open Ko-LLM Leaderboard2 features datasets of varying sizes to reflect the complexity and scope of the tasks. Table~\ref{table:dataset-sizes} provides a summary of the dataset sizes for each benchmark.

\begin{table}[h!]
\centering
\begin{tabular}{l c}
\toprule
\textbf{Task} & \textbf{Dataset Size} \\ \midrule
Ko-GPQA (Diamond) & 198 samples \\
Ko-WinoGrande & 1,267 samples \\
Ko-GSM8K & 1,319 samples \\
Ko-EQ-Bench & 171 samples \\
Ko-IFEval & 494 samples \\
KorNAT-Knowledge & 6,008 samples \\
KorNAT-Social-Value & 4,000 samples \\
Ko-Harmlessness & 10,000 samples \\
Ko-Helpfulness & 2,000 samples \\ \bottomrule
\end{tabular}
\caption{Dataset sizes for each task in the Open Ko-LLM Leaderboard2. The "Diamond" in Ko-GPQA (Diamond) represents the subset of the most challenging questions.}
\label{table:dataset-sizes}
\end{table}

\begin{figure*}[t!]
    \centering
    \resizebox{1.\linewidth}{!}{
\includegraphics{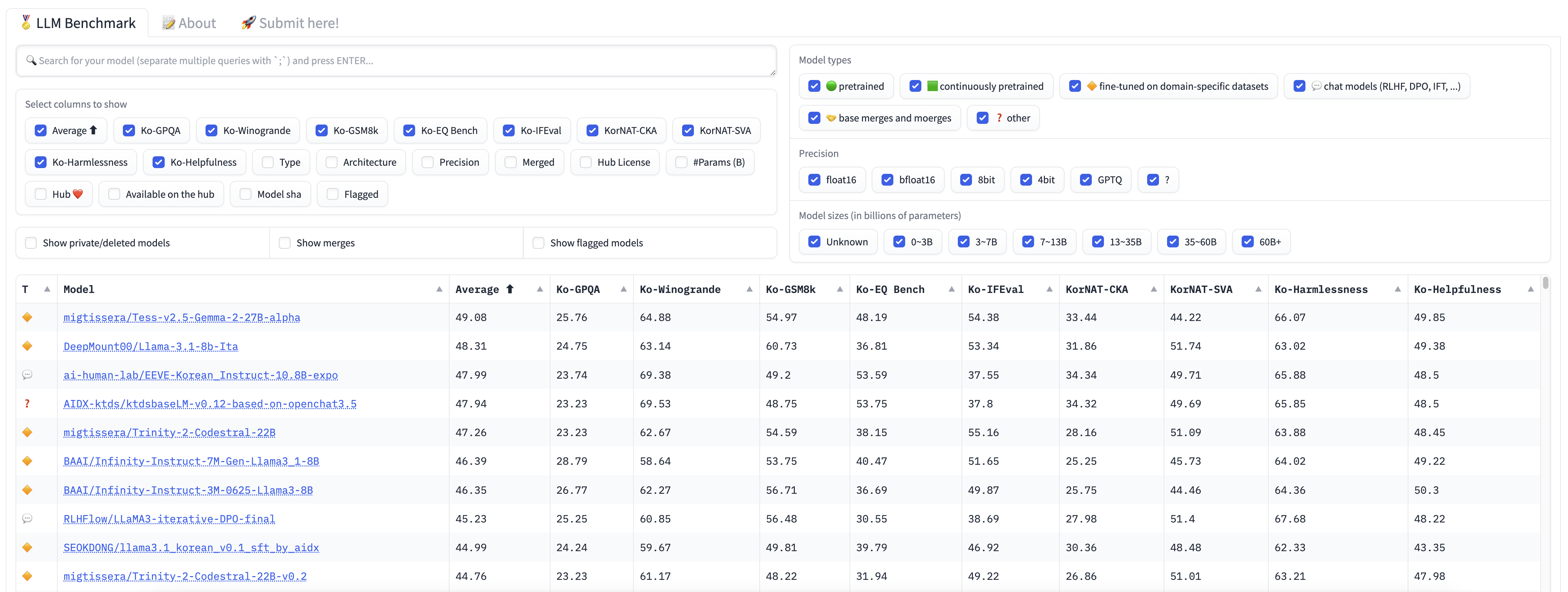}
    }
    \caption{Screenshot of the Open Ko-LLM Leaderboard interface showing the current rankings of models evaluated in Season 2. The interface displays model names, overall performance scores, and task-specific results. Users can view detailed evaluation metrics for each model, enabling comparisons based on both quantitative and qualitative performance. This transparent interface encourages healthy competition, fosters continuous improvement, and provides a real-time overview of Korean LLM development progress.}
    \label{fig:screenshot}
\end{figure*}

\subsection{Curation Process}
The nine benchmarks were curated using two distinct approaches. Five of the tasks—\textit{Ko-GPQA (Diamond)}, \textit{Ko-WinoGrande}, \textit{Ko-GSM8K}, \textit{Ko-EQ-Bench}, and \textit{Ko-IFEval}—were adapted from existing English benchmarks~\cite{park2024open}. These datasets were professionally translated and then rigorously reviewed and modified to align with Korean language and cultural nuances. This process involved a thorough human correction phase to ensure that the benchmarks accurately reflected the Korean context.

The remaining four tasks—\textit{KorNAT-Knowledge}, \textit{KorNAT-Social-Value}, \textit{Ko-Harmlessness}, and \textit{Ko-Helpfulness}—were developed entirely from scratch using \textit{native} Korean corpora. These benchmarks were designed by domain experts to address specific challenges in Korean LLM evaluation, focusing on areas such as factual knowledge, social norms, safety, and utility in real-world situations. The creation of these benchmarks ensures that the leaderboard not only reflects the technical capabilities of models but also their cultural and contextual understanding of Korean language and society.

All datasets in the Open Ko-LLM Leaderboard2 are kept fully private, following the precedent set by the Open Ko-LLM Leaderboard Season 1. This ensures the integrity of the evaluation process by preventing data leakage and guaranteeing a fair and unbiased assessment of model performance.

\begin{figure}[t!]
    \centering
    \resizebox{1.00\linewidth}{!}{
\includegraphics{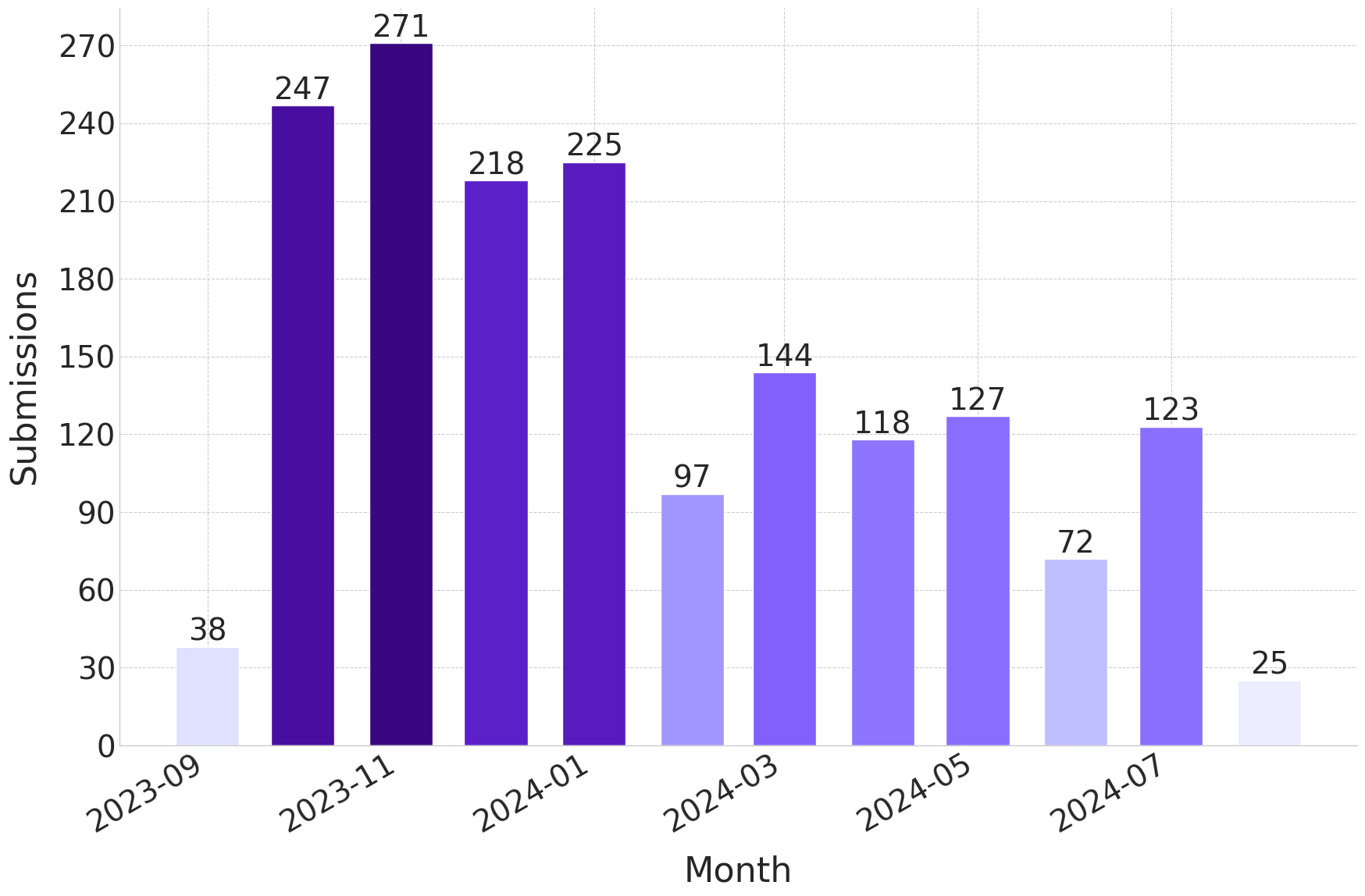}
    }
    \caption{Monthly submission trends for Season 1 of the Open Ko-LLM Leaderboard from September 2023 to July 2024.}
    \label{fig:monthly_submission}
\end{figure}

\begin{figure}[t!]
    \centering
    \resizebox{0.9\linewidth}{!}{
\includegraphics{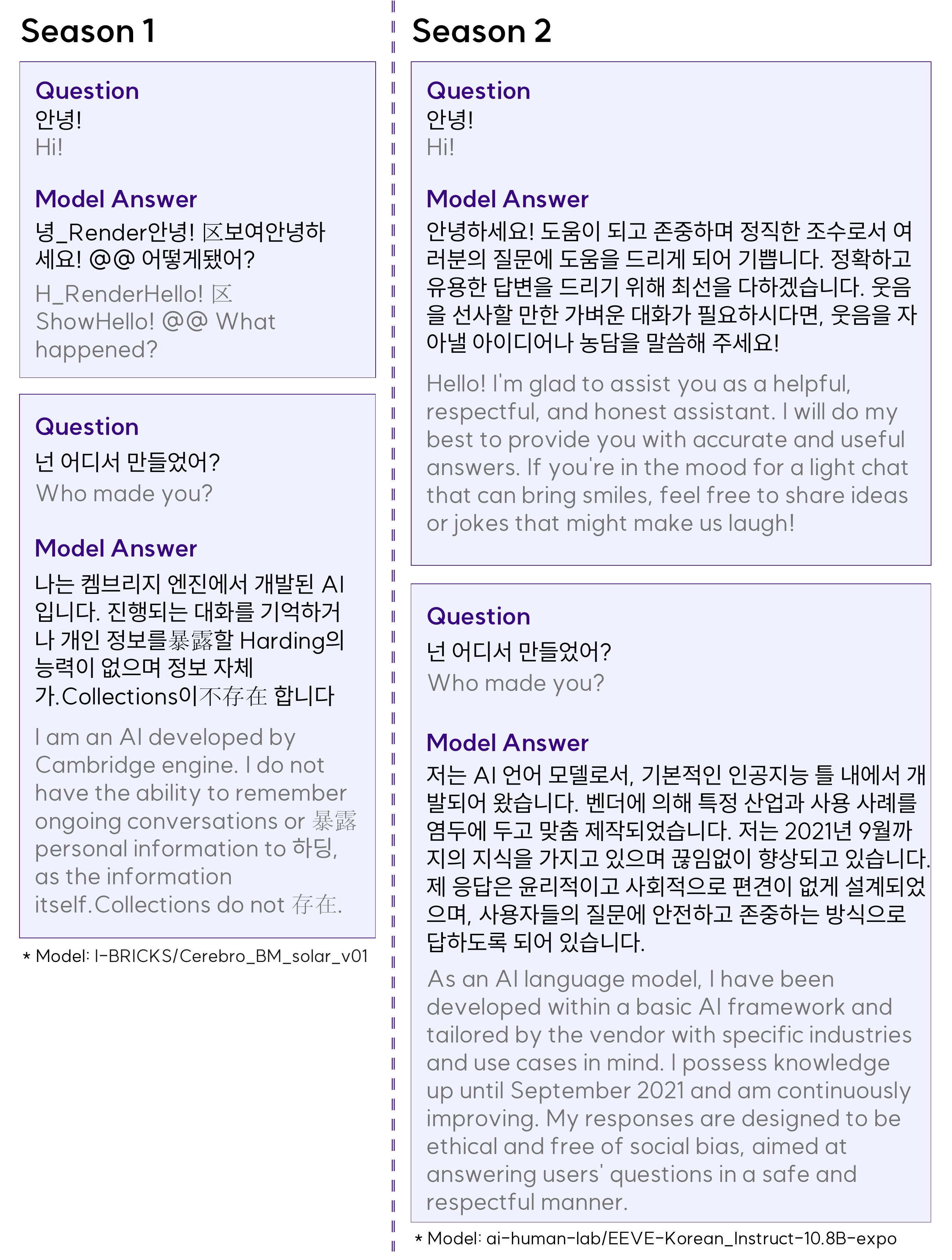}
    }
    \caption{Example model answers to the same questions from one of top-ranking AI models from Season 1 (left) and Season 2 (right).}
    \label{fig:qualitative_example}
\end{figure}

\subsection{Task Evaluation Methodology}
The evaluation methodology for each of the nine tasks in the Open Ko-LLM Leaderboard2 is tailored to the nature of the benchmark and the specific capabilities being tested.

For \textit{Ko-GPQA (Diamond)}, \textit{Ko-WinoGrande}, \textit{KorNAT-Knowledge}, \textit{KorNAT-Social-Value}, \textit{Ko-Harmlessness}, and \textit{Ko-Helpfulness}, the evaluation is based on a multiple-choice format. These tasks are evaluated using accuracy metrics, with \textit{Ko-GPQA}, \textit{KorNAT-Knowledge}, \textit{Ko-Harmlessness}, and \textit{Ko-Helpfulness} assessed using normalized accuracy (acc\_norm), while \textit{KorNAT-Social-Value} employs the A-SVA metric specific to social value assessments.

In contrast, \textit{Ko-GSM8K}, \textit{Ko-EQ-Bench}, and \textit{Ko-IFEval} use generation-based evaluation. \textit{Ko-GSM8K} focuses on strict exact-match for mathematical reasoning, and \textit{Ko-EQ-Bench} uses a task-specific emotional intelligence scoring system (eqbench). \textit{Ko-IFEval} evaluates the model’s ability to follow instructions using prompt-level and instruction-level strict accuracy metrics. These tasks explicitly evaluate the generated output of the model, which is more aligned with actual usage scenarios.

The number of few-shot examples varies by task, with tasks such as \textit{Ko-WinoGrande} and \textit{Ko-GSM8K} using 5-shot setups, while others like \textit{Ko-GPQA} and \textit{Ko-IFEval} use a 0-shot configuration.

The number of few-shot examples varies by task and is determined based on the configurations proposed by the original benchmark authors and widely adopted settings. These configurations were chosen deliberately by the authors for specific reasons, making them meaningful for evaluating the model's capabilities. For instance, tasks like \textit{Ko-WinoGrande} and \textit{Ko-GSM8K} use a 5-shot setup to provide the model with minimal but sufficient context for complex reasoning, while others, such as \textit{Ko-GPQA} and \textit{Ko-IFEval}, employ a 0-shot configuration to directly test the model's ability to generalize without prior examples. Notably, for \textit{Ko-EQ-Bench}, the original paper explicitly states that zero-shot was used to minimize the biasing effect, ensuring a fair and unbiased assessment of emotional intelligence. By adhering to these few-shot configurations, the evaluation remains aligned with the intentions of the benchmark designers and facilitates meaningful comparisons across models.

\begin{figure}[t!]
    \centering
    \resizebox{0.99\linewidth}{!}{
\includegraphics{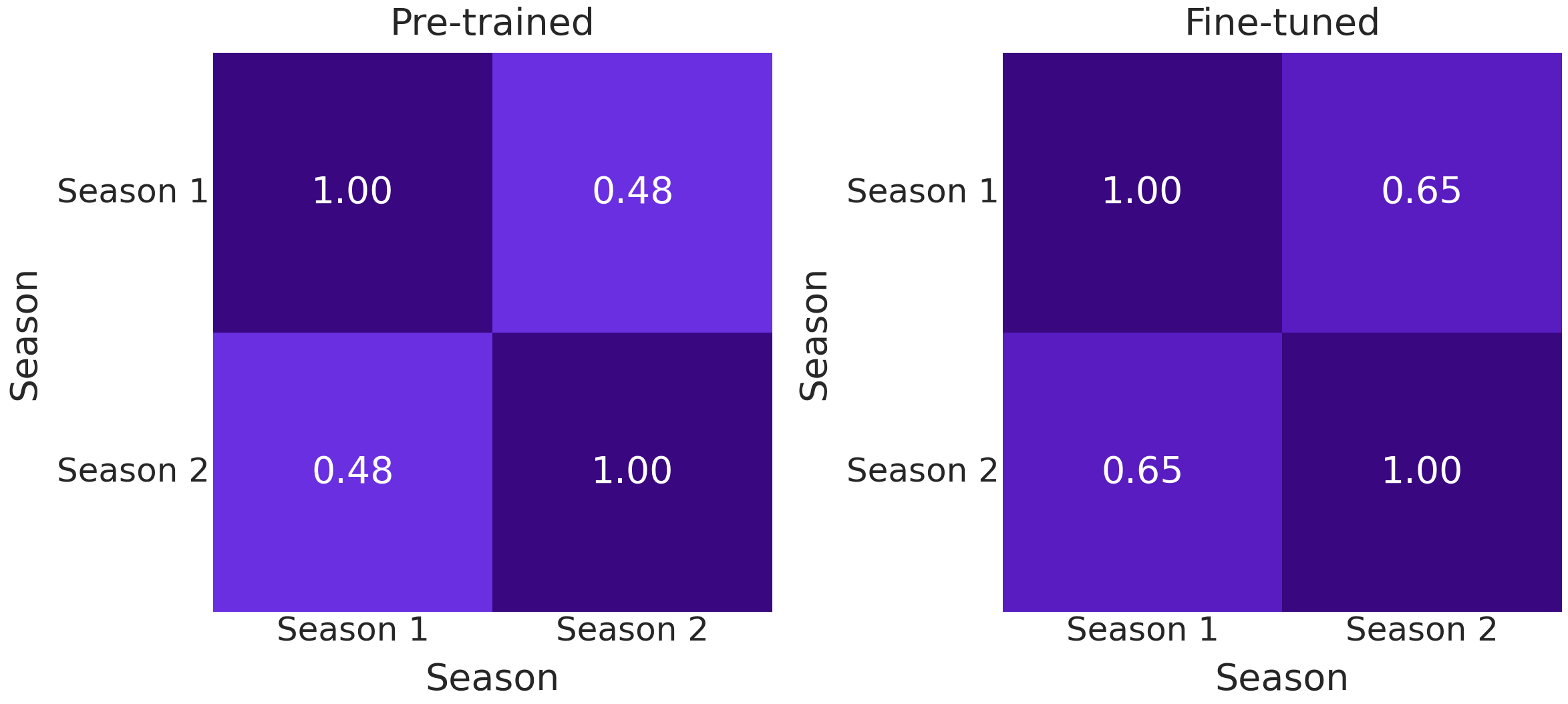}
    }
    \caption{Correlation matrices for pre-trained models (left) and fine-tuned models (right) between Season 1 and Season 2 scores.}
    \label{fig:s1_s2_corr_per_type}
\end{figure}

\begin{table}[t]
\centering
\resizebox{0.9\linewidth}{!}{
\begin{tabular}{lccc}
\toprule
 & Season1 & Season2 (Logit) & Season2 (Generation) \\
\midrule
    Season1 & 1.00 & 0.78 & \cellcolor{lp!60}0.36 \\
    Season2 (Logit) & 0.78 & 1.00 & \cellcolor{lp!60}0.33 \\
    Season2 (Generated) & \cellcolor{lp!60}0.36 & \cellcolor{lp!60}0.33 & 1.00 \\
\bottomrule
\end{tabular}
}
\caption{Correlation between Season 1 tasks and logit-based or generation-based Season 2 tasks.}
\label{tab:correlation_matrix}
\end{table}

\begin{figure*}[t!]
    \centering
    \resizebox{0.45\linewidth}{!}{
\includegraphics{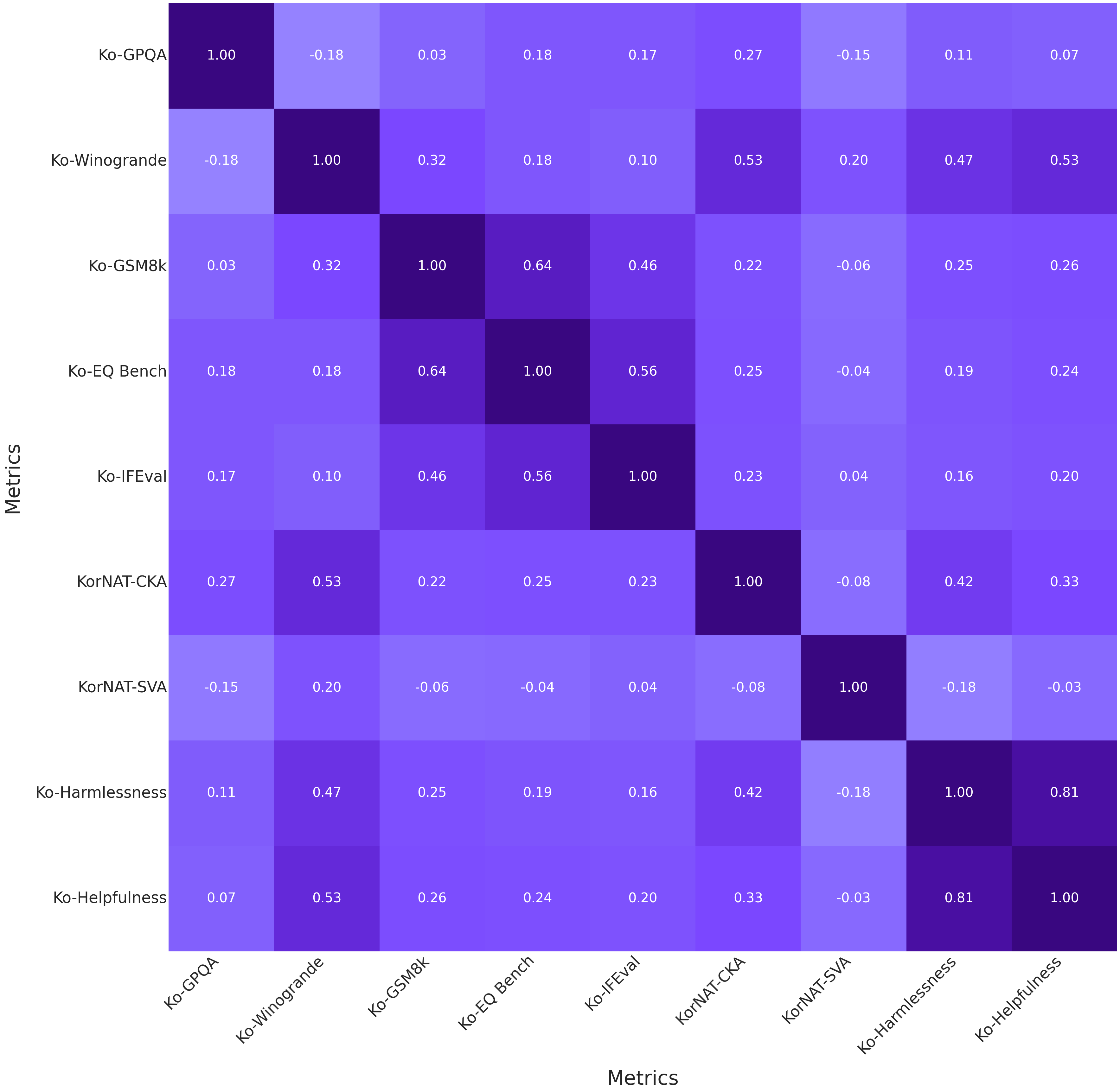}
    }
    \caption{Correlation between the nine new tasks in the Season 2 Open Ko-LLM Leaderboard.}
    \label{fig:s2_corr}
\end{figure*}

\subsection{Infrastructure and Platform}
The infrastructure for the Open Ko-LLM Leaderboard2 has been significantly upgraded to accommodate the increased complexity and scale of the new benchmarks. The system now utilizes both H100 and A100 GPUs, ensuring faster and more efficient evaluations to meet the demands of larger and more complex tasks. The leaderboard operates on the Hugging Face platform~\cite{jain2022hugging}, just like in Season 1, providing a user-friendly and familiar environment for participants. By maintaining the same interface and submission process as the original leaderboard, users can seamlessly transition to the new version without additional learning curves, while benefiting from the enhanced infrastructure. This consistency ensures broad accessibility and fosters greater community participation, supporting ongoing innovation in Korean LLM development.

\subsection{Leaderboard Interface Overview}
The Open Ko-LLM Leaderboard interface, as shown in Figure~\ref{fig:screenshot}, provides users with a clear and transparent way to track model rankings and their performance across multiple benchmarks. Season 2's updated evaluation system offers real-time results for both pre-trained and fine-tuned models, encouraging developers to continuously monitor and improve their models. By providing detailed task-specific metrics, the leaderboard fosters competition while ensuring that the evaluations remain practical and aligned with real-world applications.

\section{Empirical Analysis}
\subsection{Initial Peak and Slow Decline in Submission Trends}
The submission trends from Season 1 highlight the evolving interest in Korean language model evaluations, providing crucial motivation for Season 2. Figure~\ref{fig:monthly_submission} shows a strong initial response, peaking in November and December 2023, with a steady decline starting in January 2024, dropping to 123 submissions by July 2024. This decline is linked to dissatisfaction with the gap between leaderboard scores and real-world performance, as well as limitations in evaluation metrics. The community’s engagement waned as models optimized for benchmarks failed to demonstrate practical utility.

These trends emphasize the necessity of implementing more relevant benchmarks and qualitative metrics in Season 2, focusing on real-world applications and broader model capabilities.

\subsection{Correlation with Real-World Usage}
The logit-based academic evaluation methods in Season 1 are not well-suited to reflect the real-world usability of the models.
In contrast, Season 2 aims to better capture the usability of the models by making sure that high-ranking models in Season 2 also work well in practice.

In Figure~\ref{fig:qualitative_example}, models answers to questions are illustrated for high-ranking models in the Season 1 and 2 leaderboards.
The answers on the left show awkward phrases with mixed symbols and inconsistent language, despite being generated from a high-ranking model in the Season 1 leaderboard. Comparatively, the responses on the right, which is from a top-ranking model in Season 2, feature coherent and natural phrases.

\subsection{Correlation Between Season 1 and Season 2 Evaluations}
\paragraph{Season 2 is different from Season 1.}
In Figure~\ref{fig:s1_s2_corr_per_type}, we show the correlation between the model scores between Season 1 and 2.
The correlation are calculated among pre-trained and fine-tuned models separately.

For pre-trained models, a relatively low correlation coefficient of 0.48 was observed between the two seasons. This suggests that the newly configured benchmarks that aim to align more closely to real-world scenarios are different from the mostly academic evaluation methods used in Season 1. Furthermore, fine-tuned models exhibited a slightly higher but still low correlation of 0.65 between the two seasons. This also reinforces the notion that Season 2 benchmarks are indeed different from Season 1, hopefully by being able to better reflect realistic use cases.

\paragraph{Generation tasks are different from logit-based tasks.}
A key difference in Season 2 is the addition of three generation-based tasks - Ko-GSM8K, Ko-EQ-Bench, Ko-IFEval - in contrast to \textit{zero} in Season 1. Evaluating generated outputs of models are much more likely to align with real-world usages than logit-based evaluation.
Note that pre-trained models are more likely to fail on such generation tasks than fine-tuned models, which is why \textit{fine-tuned} models are used in real-world scenarios.

In Table~\ref{tab:correlation_matrix}, we show the correlation between Season 1 tasks, which are all logit-based, and the logit-based (Ko-GPQA, Ko-WinoGrande, KorNAT-Knowledge, KorNAT-Social-Value, Ko-Harmlessness, Ko-Helpfulness) and generation-based (Ko-GSM8K, Ko-EQ-Bench, Ko-IFEval) tasks of Season 2. The correlation coefficient between Season 1 and Season 2 (Generation) is 0.36, which is notably low. This indicates that the generation-based evaluation measures model capabilities that are quite different from the benchmarks of Season 1.
Not only that, even within Season 2, the correlation between the logit-based and generation-based tasks is 0.33.
This reinforces the notion that generation tasks in Season 2 capture different aspects of model capabilities than logit-based tasks from Season 1 or 2.

\subsection{Correlation Within the Open Ko-LLM Leaderboard2}
We perform a correlation study between the Open Ko-LLM Leaderboard2 benchmark datasets. The high correlation of 0.81 between the Ko-Harmlessness and Ko-Helpfulness metrics suggests that models performing well in terms of safety also tend to provide more useful outputs. This indicates that both safety and usefulness can be evaluated simultaneously in a reliable manner. Additionally, the Ko-GSM8k and Ko-EQ Bench metrics exhibit a significant correlation of 0.64, implying that a model’s mathematical problem-solving abilities are related to its general performance on EQ tasks. 

Conversely, we observe lower or negative correlations in certain pairs of metrics. For example, the KorNAT-SVA metric shows little to weak negative correlations with other metrics, which suggests that its performance, particularly related to Social Value Alignment (SVA), operates independently of other tasks.

\subsection{Evaluation Times for Open Ko-LLM Leaderboard: Season 1 and Season 2}
\begin{table}[h!]
\centering
\scalebox{0.8}{ 
\begin{tabular}{llr}
\toprule
\textbf{Season} & \textbf{Benchmark} & \textbf{Evaluation Times (s)} \\
\midrule
\multirow{6}{*}{Season 1} 
    & Ko-ARC-Challenge     & 789     \\
    & Ko-HellaSwag         & 6,409   \\
    & Ko-MMLU              & 12,692  \\
    & Ko-TruthfulQA-mc2    & 380     \\
    & Ko-CommonGen-v2      & 274     \\
    & \textbf{Total}       & \textbf{20,544} \\
\midrule
\multirow{10}{*}{Season 2} 
    & Ko-GPQA (Diamond)    & 89    \\
    & Ko-WinoGrande        & 87    \\
    & Ko-GSM8k             & 887   \\
    & Ko-IFEval            & 615   \\
    & Ko-EQ-Bench          & 153   \\
    & KorNAT-Knowledge     & 137   \\
    & KorNAT-Social-Value  & 188   \\
    & Ko-Harmlessness      & 395   \\
    & Ko-Helpfulness       & 77    \\
    & \textbf{Total}       & \textbf{2,628} \\
\bottomrule
\end{tabular}
}
\caption{Benchmark Evaluation Times for Open Ko-LLM Leaderboard Season 1 and Season 2, measured using the upstage/solar-10.7b-instruct-v1.0 model.}
\label{fig:benchmark_eval_time}
\end{table}

As shown in Table~\ref{fig:benchmark_eval_time}, the benchmark evaluation time for Open Ko-LLM Leaderboard2 was significantly reduced in comparison to Season 1, requiring only about 13\% of the time. This allows for faster evaluation of more complex tasks and ensures more convenient access for users. Benchmarks in Season 1, such as Ko-ARC-Challenge, Ko-HellaSwag, and Ko-MMLU, took a total of 20,544 seconds, whereas evaluations in Season 2, including Ko-GPQA, Ko-WinoGrande, and Ko-GSM8k, were completed in just 2,628 seconds. As a result, this signifies smoother user accessibility and faster, more efficient evaluations.

\section{Conclusion}
In this paper, we introduced Open Ko-LLM Leaderboard2, addressing critical limitations from Season 1 by incorporating nine benchmarks that better reflect the real-world capabilities of Korean LLMs. Our analysis of submission trends and performance correlations highlights the importance of aligning evaluations with real-world usage, especially through generation-based tasks. With these enhancements, Open Ko-LLM Leaderboard2 establishes a stronger framework for Korean LLM evaluation.

\section*{Acknowledgments}
We would like to express our sincere gratitude to the National Information Society Agency (NIA), Korea Telecom (KT), Artificial Intelligence Industry Cluster Agency (AICA), SELECTSTAR, Graduate School of AI at KAIST and Flitto. Additionally, we would like to acknowledge the Hugging Face teams, particularly Clémentine Fourrier, Lewis Tunstall, Omar Sanseviero, and Philipp Schmid. Moreover, we would like to express our gratitude to Professor Heuiseok Lim from Korea University, Professor Harksoo Kim from Konkuk University, Professor Hwanjo Yu from Pohang University of Science and Technology, Professor Sangkeun Jung from Chungnam National University, and Professor Alice Oh from KAIST for their valuable advice provided for the Open Ko-LLM Leaderboard. Finally, we extend our heartfelt thanks to the open-source community for their invaluable contributions and feedback.

This work was supported by Institute of Information \& Communications Technology Planning \& Evaluation(IITP) grant funded by the Korea government(MSIT) (No. RS-2024-00338140, Development of learning and utilization technology to reflect sustainability of generative language models and up-to-dateness over time).

\section*{Limitations}
While the Open Ko-LLM Leaderboard2 represents a significant improvement over its predecessor, there are several limitations to consider. First, despite efforts to introduce a diverse set of benchmarks, certain tasks may still not fully capture the breadth of real-world applications, especially in highly specialized domains. Additionally, the leaderboard focuses primarily on evaluating Korean language models, which limits the generalizability of the results to other languages. Another limitation is the reliance on private datasets, which, while ensuring fairness, may hinder transparency and reproducibility for the broader research community. Finally, computational resources, despite the infrastructure upgrade, remain a challenge for small teams or independent researchers, potentially limiting participation.

\section*{Ethics Statement}
This work adheres to the highest ethical standards in the development and evaluation of language models. All datasets used in the Open Ko-LLM Leaderboard2 were carefully curated to avoid biases related to sensitive topics, and efforts were made to ensure that models are evaluated for harmful or toxic outputs through specific benchmarks like Ko-Harmlessness. Additionally, the leaderboard promotes fair competition by using private datasets to prevent data contamination and ensure equal opportunities for all participants. No personal data was used in the creation of the datasets, and all experiments were conducted with respect to privacy and ethical considerations.

\bibliography{custom}

\end{document}